\documentclass[letterpaper, 10 pt, conference]{ieeeconf}  

\IEEEoverridecommandlockouts                              
\usepackage{graphics}
\usepackage[caption=false]{subfig}
\usepackage{graphicx}
\usepackage{balance}

\usepackage{cite}

\usepackage{amsmath}
\usepackage{amssymb}  

\usepackage{amsthm}  
\usepackage{algorithm}
\usepackage{algpseudocode}
\usepackage{bm}
\algdef{SE}[DOWHILE]{Do}{doWhile}{\algorithmicdo}[1]{\algorithmicwhile\ #1}%

\usepackage{lipsum}
\usepackage{todonotes}
\usepackage{balance}

\usepackage{mathtools}

\DeclarePairedDelimiterX{\infdivx}[2]{(}{)}{%
  #1\;\delimsize\|\;#2%
}

\usepackage{siunitx}

\DeclareMathOperator*{\argmin}{arg\,min}

\usepackage{amsthm}

\theoremstyle{definition}

\newtheorem{definition}{Definition}

\title{\LARGE \bf
Toward Optimal FDM Toolpath Planning with Monte Carlo Tree Search
}

\author{Chanyeol~Yoo$^1$,
        Samuel~Lensgraf$^2$,
        Robert~Fitch$^1$,
        Lee~M.~Clemon$^1$,
        and Ramgopal~Mettu$^3$
    \thanks{This research is supported in part by the University of Technology Sydney, Dartmouth College, and Tulane University.}
	\thanks{$^1$School of Mechanical and Mechatronic Engineering, University of Technology Sydney, NSW 2006, Australia {\tt\footnotesize \{Chanyeol.Yoo, Lee.Clemon, Robert.Fitch\}@uts.edu.au}}
	\thanks{$^2$Department of Computer Science, Dartmouth College, Hanover, NH 03755, USA {\tt\footnotesize Samuel.E.Lensgraf.GR@dartmouth.edu}}
	\thanks{$^3$Department of Computer Science, Tulane University, New Orleans, LA 70118, USA {\tt\footnotesize rmettu@tulane.edu}}
    }

\begin{document}

\maketitle
\thispagestyle{empty}
\pagestyle{empty}

\begin{abstract}
The most widely used methods for toolpath planning in fused deposition 3D printing
slice the input model into successive 2D layers in order to construct the toolpath. Unfortunately slicing-based methods can incur a substantial amount of wasted motion (i.e., the extruder is moving while not printing), particularly when features of the model are spatially separated. In recent years we have introduced a new paradigm that characterizes the space of feasible toolpaths using a dependency graph on the input model, along with several algorithms to search this space for toolpaths that optimize objective functions such as wasted motion or print time. A natural question that arises is, under what circumstances can we efficiently compute an optimal toolpath? In this paper, we give an algorithm for computing fused deposition modeling (FDM) toolpaths that utilizes Monte Carlo Tree Search (MCTS), a powerful general-purpose method for navigating large search spaces that is guaranteed to converge to the optimal solution. Under reasonable assumptions on printer geometry that allow us to compress the dependency graph, our MCTS-based algorithm converges to find the optimal toolpath. We validate our algorithm on a dataset of 75 models and show it performs on par with our previous best local search-based algorithm in terms of toolpath quality. In prior work we speculated that the performance of local search was near optimal, and we examine in detail the properties of the models and MCTS executions that lead to better or worse results than local search. 
\end{abstract}

\section{INTRODUCTION}

Making 3D printing more efficient is a pressing need in the area of additive manufacturing \cite{huang_additive_2015,ngo_additive_2018}. The current sluggish throughput of 3D printers keeps costs above traditional manufacturing lines and prevents adoption in medium and high volume manufacturing. Fabrication speed has long been and remains a major limiting factor for commercial adoption of additive manufacturing processes \cite{kruth_progress_1998,huang_additive_2015,ngo_additive_2018}. While adoption of 3D printing in medical and aerospace industries is driven by the low volume and high value products, consumer products and medium to high volume production depend on rapid fabrication to meet demand and keep costs competitive. Thus, tackling the speed of fabrication in 3D printing is a sector-wide challenge. Current deposition planning methods in 3D printing depend on 2D layering and complete each entire layer sequentially, resulting in very long build times and thus high cost per product. A major contributor to this additional cost is wasted motion hoping around within each layer. In this work we greatly reduce this wasted motion by reexamining the entire deposition planning order.

In prior work we have introduced a framework for optimizing toolpaths with respect to both wasted motion \cite{LensgrafMettuICRA2016,LensgrafMettuICRA2017} as well as overall time \cite{LensgrafMettuIROS18}. We introduced the notion of a dependency graph of a model to be fabricated, which is defined by the geometric properties of the extruder head. Conceptually the dependency graph constrains the order in which model features must be fabricated and thus allows us to define the space of toolpaths which would feasibly construct the input model on the given printer geometry. We also gave simple heuristic algorithms based on greedy and local search to search over the space of toolpaths. These algorithms all demonstrated that significant gains could be made in toolpath efficiency, especially with respect to wasted motion, if we were not constrained to fabricate layer-by-layer. A natural question that arises from these approaches is, what is the \textit{optimal} toolpath that can be computed for a given input model and printer geometry? A straightforward, and discouraging, observation is that planning a single layer of a fused deposition modeling (FDM) toolpath is easily seen to reduce to TSP. Thus the scale of planning an entire toolpath, or even estimating the optimal cost, using approximation algorithms for TSP or methods such as the Held-Karp relaxation to obtain LP lower bounds is intractable. In \cite{LensgrafMettuICRA2017} we explored LP lower bounds to characterize the performance of local search, but even then were only able to demonstrate optimality for a few, very small models.

Our main contribution in this paper is to develop the first Monte Carlo Tree Search (MCTS) approach to toolpath planning. MCTS can be seen as a stochastic branch and bound approach in which a probabilistic tree search is used to balance exploration and exploitation in the search space. MCTS is a flexible, anytime approach to optimization problems, and is guaranteed to converge to an optimal solution. To our knowledge, this is the first algorithm for toolpath planning with any guarantees on global optimality. In recent years MCTS has been shown to be effective for difficult problems in a number of application areas in robotics. Of particular note is the  Dec-MCTS~\cite{best2019decmcts} framework, which allows MCTS to be solve optimization problems in an asynchronous, coordinated fashion. By applying the Dec-MCTS framework, the algorithm then yields an efficient approach to multi-extruder toolpath planning with guaranteed convergence, which to our knowledge does not yet exist. In order to make our algorithm efficient, we also introduce a novel clustering algorithm on the dependency graph for the input model. 
Using a dataset of 75 models, we show that our MCTS-based approach achieves a substantial reduction in wasted motion over the layer-by-layer toolpath generated by Sli3r~\cite{slic3r}. The performance of MCTS is interestingly quite similar to that of our existing local search toolpath planner, but allows us to more carefully examine the difficulty of toolpath planning for certain types of models. We study the empirical performance of MCTS and attempt to gather insights about the optimality of our toolpaths. In particular, we have strong evidence of convergence for certain models, and we provide criteria for assessing optimality from our observations.

\section{RELATED WORK}

Previous research on toolpath planning and slicing of 3D printed parts as focused on examining the resolution of the build relative to the CAD model \cite{pandey_real_2003,dolenc_slicing_1994}. This area of work led to varying layer thickness to increase fabrication speed, but achieve a specified surface resolution \cite{dolenc_slicing_1994,kulkarni_accurate_1996,pandey_real_2003}. Others have focused on adjusting the orientation of the layering through transitions in multi-axis machines \cite{isa_five-axis_2019}, non-planar layers \cite{isa_five-axis_2019,micali_fully_2016,ahlers_3d_2018}, reducing support structures \cite{paul_optimization_2015}, or segmenting areas within a layer \cite{jin_optimization_2014}. These works do not consider the segments within each layer as part of the entire build plan, but focus on each layer individually. Prior work that focuses on path planning is primarily concerned with increasing the speed of the mechanisms or material reaction rates \cite{kruth_progress_1998,go_rate_2017}, or varying the thickness of each slice \cite{ma_adaptive_1999} or layering non-planar surfaces \cite{micali_fully_2016,ahlers_3d_2018}. In contrast to our prior work~\cite{LensgrafMettuICRA2016,LensgrafMettuICRA2017,LensgrafMettuIROS18}, the former approaches depend on the same sequential and ordered layering as the original 2.5D methods. Thus, the approach in this work and our related prior work represents a fundamental shift away from individual layer-planning towards a generalized concept of build-planning for additive manufacturing. 

MCTS is a sequential Monte Carlo approach for efficient planning in large state spaces~\cite{browne2012survey,kocsis06a}. It has been used in motion planning for robot systems in a variety of tasks such as active information gathering~\cite{tim_AURO2017,hefferan2016patrolling,kartal2015patrol,nguyen2015glider}. DEC-MCTS is a decentralised form of this algorithm~\cite{best2019decmcts,best2018deccomms} that has been used for motion planning in multi-robot systems such as a pair of robot manipulators~\cite{sukkar2019roi}.

\section{PRELIMINARIES AND PROBLEM FORMULATION}

Given an input 3D model for fabrication, as in prior work we assume that the given model has already been decomposed into individual, printable line segments. Let this set of line segments be denoted as $L  = \{\ell_1, \cdots, \ell_n\}$. Each line segment~$\ell_i$ is an ordered pair~$(\mathbf{x}_{i, 0}, \mathbf{x}_{i, 1})$, where~$\mathbf{x}_{i, \cdot} = [x, y, z]^T \in \mathbb{R}^3$. Given $L$, we define the associated set of \textit{contours} $C$ as a partition of $L$ into continuous series of line segments. Intuitively a contour is a series of line segments in the input model that could be printed in one continuous extrusion. Note that a contour could be either opened or closed sequence of line segments.

Let the distance to print a line segment~$\ell_i$ be denoted as~$p_i \in \mathbf{P}$, and the distance to move the extruder between a point $\mathbf{x}_j$ to~$\mathbf{x}_k$ as~$w_{\mathbf{x}_j, \mathbf{x}_k}$. We will write the distance required to travel between line segments $\ell_i$ and $\ell_j$ as $w_{i, j}$, where~$w_{i, j} = c(\mathbf{x}_{i, 1}, \mathbf{x}_{j, 0})$. That is, the travel distance between two line segments is simply the distance to travel from the end of the first line segment to the beginning of the second line segment. A \textit{toolpath} is simply a numbered ordering of contours~$\pi: C \rightarrow \mathbb{N}$. The \textit{cost} of a given toolpath is the sum of the distances to print line segments and to travel between line segments. 

We note that our definition of a toolpath does not take the printer geometry into consideration; indeed a low cost toolpath may actually not be feasible (i.e., correctly produces the input model) on all printing platforms. In order to model the constraints imposed on the toolpath by the geometry of the printer, in~\cite{LensgrafMettuICRA2016} we introduced the notion of a \textit{dependency graph} that is defined by the contours in the input model and printer geometry. We define the dependency graph $D = (C, E)$ such that the contours $C$ are the vertices, and a directed edge $(c_i, c_j)\in E$ if $c_i$ must be printed prior to $c_j$ in any feasible toolpath. For an edge $(c_i, c_j)\in E$, we denote~$c_i$ as the \emph{dependee} and~$c_j$ as \emph{depender} for the pair. Intuitively, all dependees must be printed before a depender can be printed.
In this paper, we will generate the dependency graph by considering the bounding volume of the extruder head, and defining the dependency graph edges $(c_i, c_j)$ by whether a $c_i$ would be unreachable after printing $c_j$ due to the extruder geometry. We note that the dependency graph formulation can be made substantially more general if we wish to take other constraints into consideration. 

With these definitions in mind, we can see that a toolpath $\pi$ that feasibly prints the input model must be a sequence of contours that satisfies: i) $\pi$ contains all contours, and ii) for any contour $c_i$ and any subsequent contour $c_j$ in $\pi$, $(c_j, c_i)\not\in D$. In other words, $\pi$ fabricates the input model without violating the geometric constraints of the extruder. The power of this view of the space of all toolpaths is that we can now define the \textit{optimal toolpath}, which is simply 
\begin{equation} \label{eqn:problem_mintotal}
    \pi^* = \argmin_{\pi} p_{i} + w_{i, j}.
\end{equation}
This is a toolpath that minimizes movement of the extruder without printing called \emph{extrusionless travel distance}. Since the sum of printing all line segments~$P$ is independent of toolpath, the problem can be re-written more simply as
\begin{equation} \label{eqn:problem_minwaste}
    \pi^* = \argmin_{\pi} w_{i, j}
    .
\end{equation}

\begin{figure}[t!]
	\centering
	\subfloat[STL model for `four nuts' model]{\includegraphics[width=0.8\linewidth]{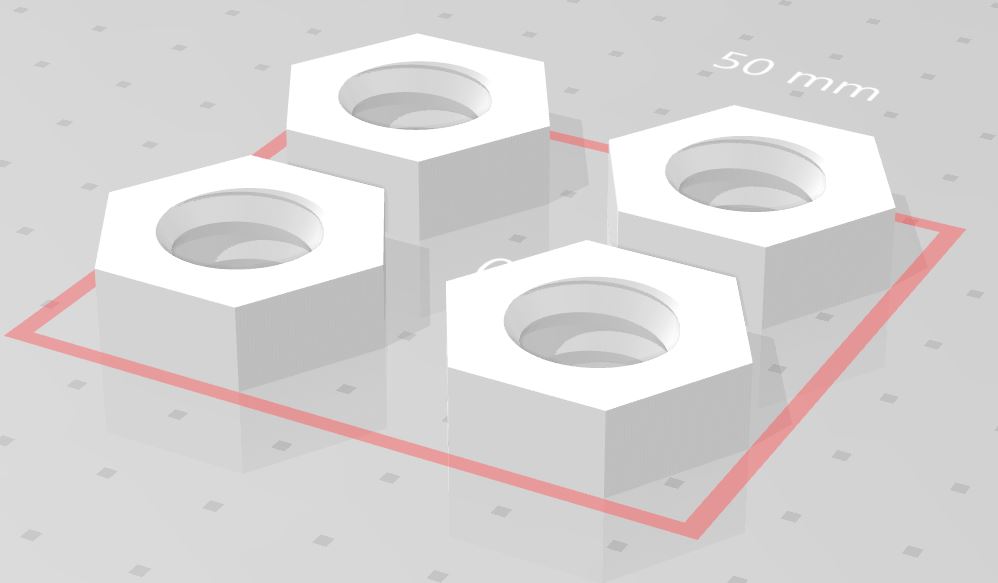}}\\
	\subfloat[Dependency graph and highly dependent subgraphs]{\includegraphics[width=0.85\linewidth]{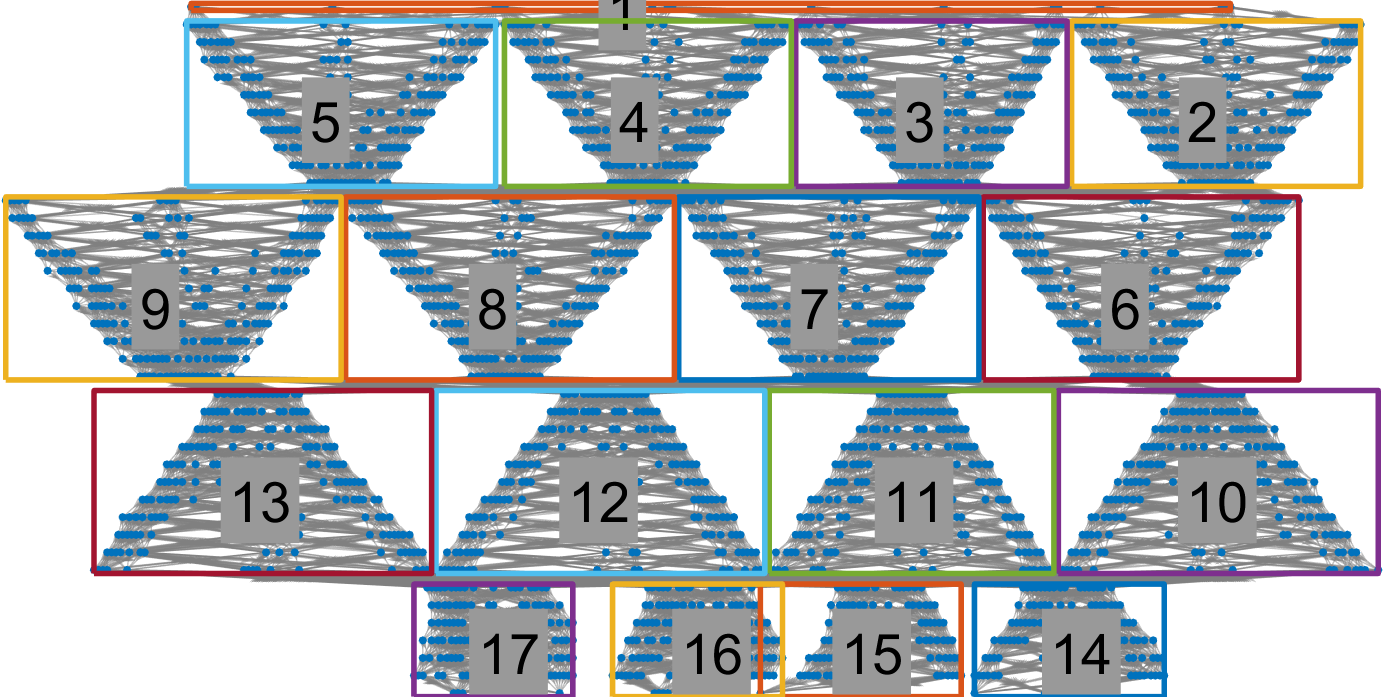} \label{fig:four_nuts_dep}}
	\\
	\subfloat[Dependency graph over highly dependent subgraphs in (b)]{\includegraphics[width=0.85\linewidth]{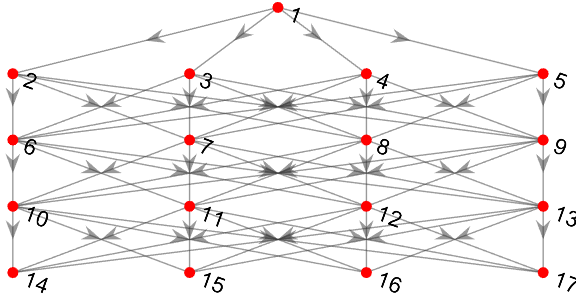} \label{fig:four_nuts_dep_clustered}}
	\caption{Example model of `four nuts' (a) image, (b) labelled dependency graph, and (c) clustered dependency graph from (b)}
	\vspace{-2ex}
	\label{fig:four_nuts}
\end{figure}

\section{MCTS-BASED TOOLPATH PLANNING} \label{sec:planning}

In this section we describe our Monte Carlo Tree Search (MCTS) toolpath planning framework. We first describe the MCTS approach, and then our adaptation of it for toolpath planning.

\subsection{Monte Carlo Tree Search} \label{sec:mcts}


Monte Carlo tree search~\cite{browne2012survey} algorithm is based on biased search algorithm for finding an optimal solution asymptotically. Starting at an initial condition, a tree grows at every iteration. The algorithm finds the next best node in a tree to expand using \emph{upper confidence bound} (UCB), where UCB balances between exploitation and exploration. Intuitively, the node with higher likelihood of finding a better solution will be selected. Once a node is selected for expansion, one or a number of complete sequence is randomly generated from the node until reaching the end (e.g., end of time horizon). We call such complete sequence a \emph{rollout}. Each random rollout is evaluated and the \emph{reward} for the rollout is backprogated up to the root node. 

Since the algorithm works by randomly generating at least one rollout at every iteration, MCTS is an \emph{anytime} algorithm, where a solution exists whenever the algorithm is stopped. This is because any rollout is a valid solution to the problem. Also, the algorithm will eventually find a global optimal solution as we repeat the process due to its random nature. Unlike existing Monte Carlo methods, MCTS finds such solution faster since it biases its search using UCB. 

In this paper, we implement MCTS to find a solution that solves for Problem~\ref{eqn:problem_minwaste} where we aim to minimise the total extrusionless travel distance of the tooltip. 
Given a set of contours~$C$ and dependency graph~$D$, a valid rollout~$\pi$ is a sequence of contours in a form
\begin{equation} \begin{split}
    &\pi = \rho_1 \rho_2 \cdots \rho_{|C|}    \\
    &\text{s.t.}~R(\rho_i) \models~\text{true}~\forall i \in [1, |C|]
    ,
\end{split} \end{equation}
where~$\rho_i \in C$ is the $i$-th contour to print and~$R(\rho_i)$ recursively evaluates whether~$\rho_i$ is printable given contours already printed. Before a contour~$\rho_i$ can be printed, all its dependees must be printed according to~$D$. 

Given a rollout sequence of contours, we compute the reward~$r$ in the form of extrusionless travel distance~$T$. Since rewards in MCTS must be a value between 0 and 1, the travel distance is normalised by~$r = \frac{\hat{t} - T}{\hat{t}}$, where~$\hat{t}$ is the upper bound for travel distance. The upperbound is calculated by multiplying the longest extrusionless travel distance within contours~$C$ by the total number of contours. Intuitively, the reward approaches zero as the travel distance approaches the upper bound and it approaches one as the distance gets closer to zero travel distance.

\subsection{Dependency Clustering} \label{sec:clustering}

\begin{figure}[tb]
	\centering
	\subfloat[]{
	    \includegraphics[height=0.4\columnwidth]{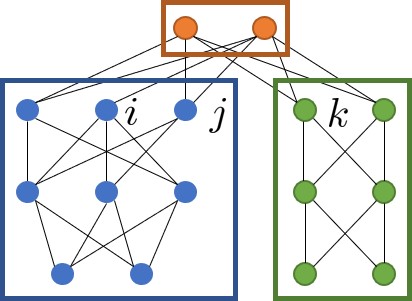}
	    \label{fig:example_hds}
	}
	\hspace{0.2cm}
	\subfloat[]{
	    \includegraphics[height=0.4\columnwidth]{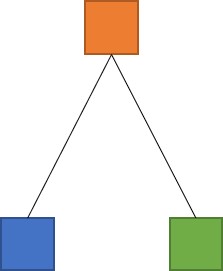}
	    \label{fig:example_clustered}
	}
	\caption{
        An example illusting clustering algorithm in Alg.~\ref{alg:clustering}. (1) 16 raw contours are clustered into three highly dependent subgraphs (HDS) as shown in (b).
	}
	\vspace{-2ex}
	\label{fig:example_clustering_overview}
\end{figure}

\begin{figure*}[t!]
	\centering
	\subfloat[`four nuts']{
	    \includegraphics[width=0.45\columnwidth]{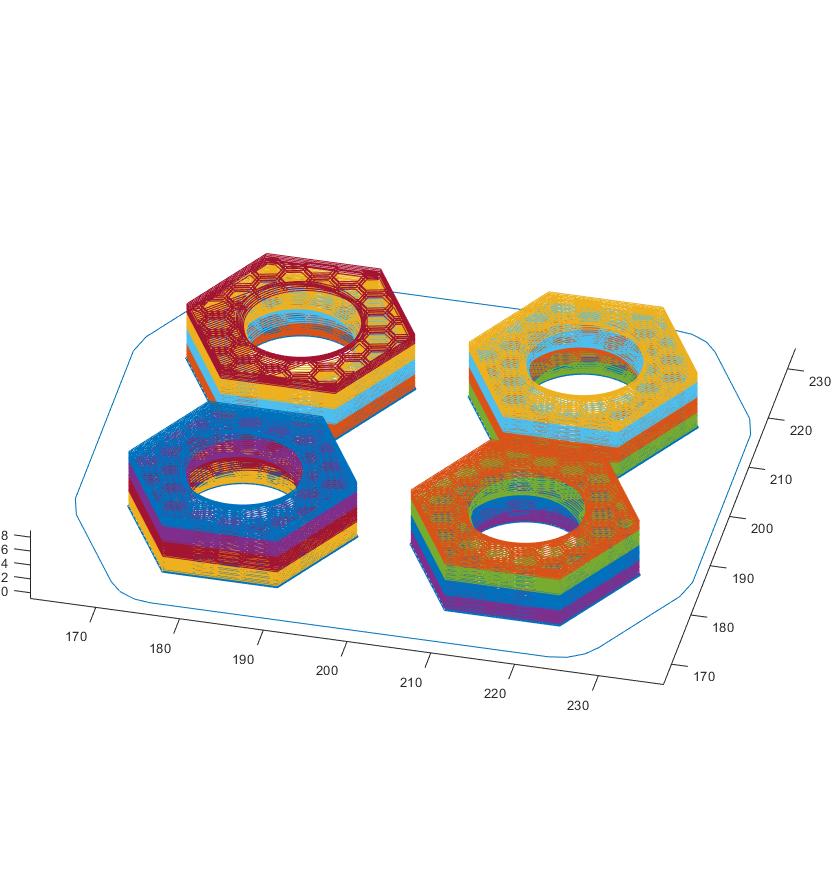} \label{fig:clustering_four_nuts}
	}
	\subfloat[`four screws']{
	    \includegraphics[width=0.45\columnwidth]{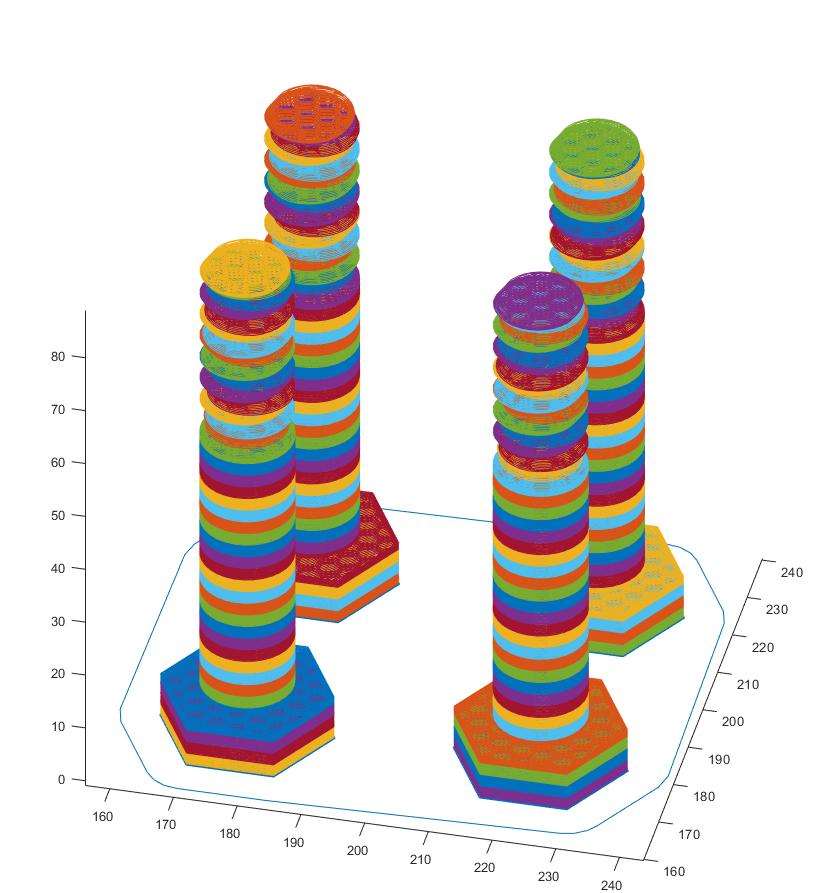} \label{fig:clustering_four_screws}
	}
	\subfloat[`twisty']{
	    \includegraphics[width=0.45\columnwidth]{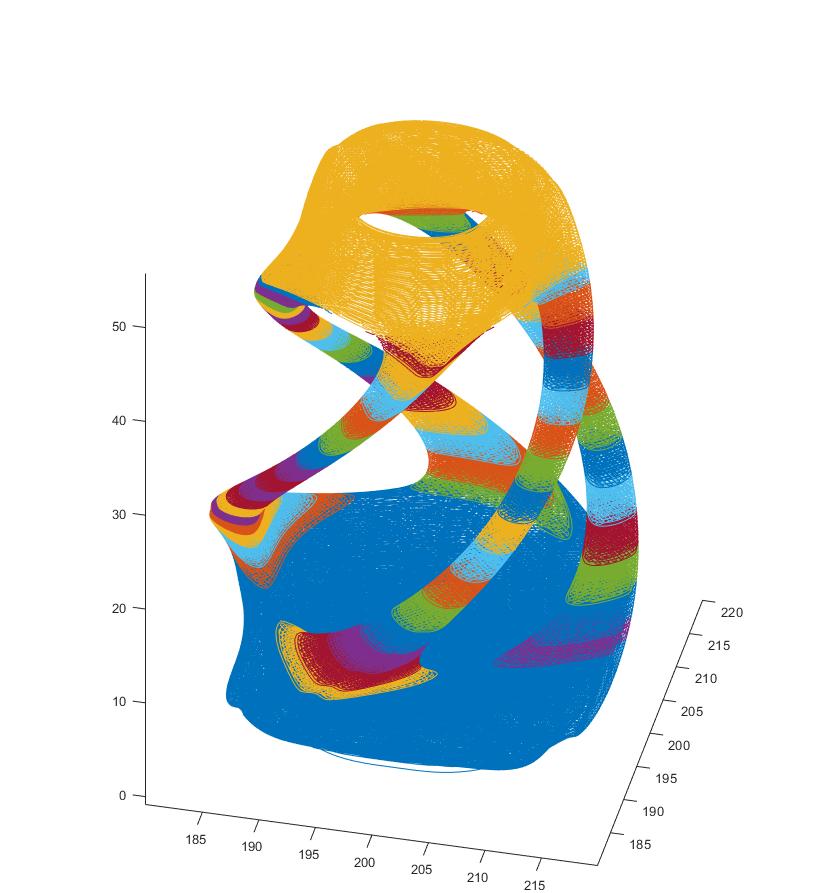} \label{fig:clustering_twisty}
	}
	\subfloat[`10-ascii']{
	    \includegraphics[width=0.45\columnwidth]{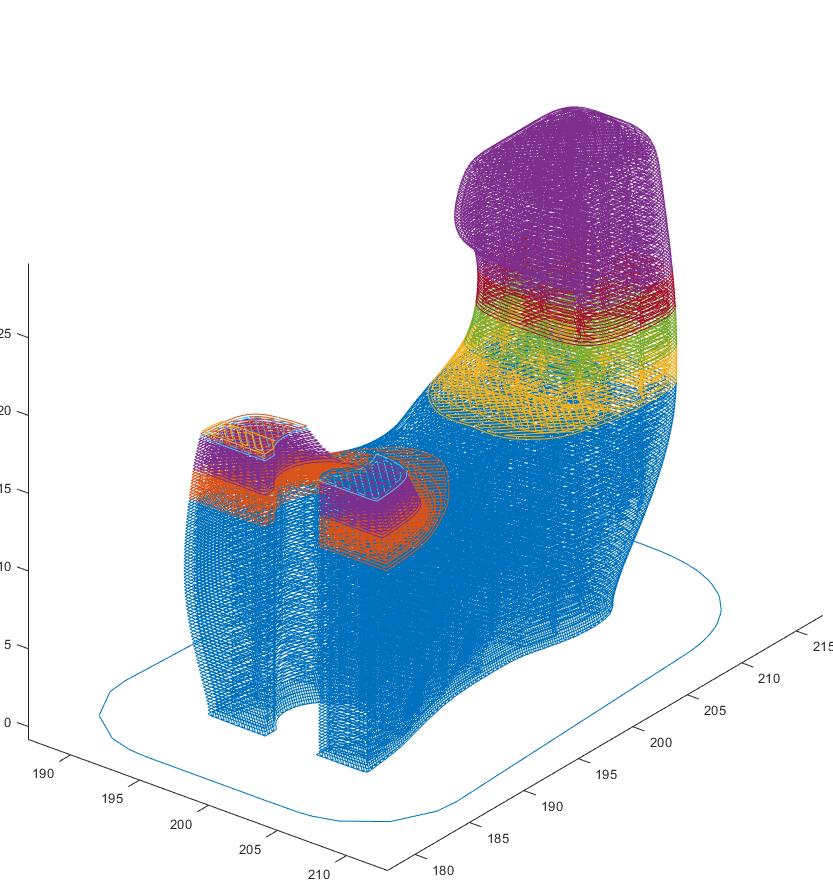} \label{fig:clustering_10-ascii}
	}
	\caption{
        Colored clusters for example parts
	}
	\vspace{-2ex}
	\label{fig:clustering_examples}
\end{figure*}

\begin{algorithm}[t]
	\caption{Dependency Clustering} \label{alg:clustering}
	\textbf{Input:}~Dependency graph~$D$   \\
	\textbf{Output:}~A set of HDS~$H$
	\begin{algorithmic}[1]
		\State $H \leftarrow \emptyset$
		\ForAll{dependency depth~$d \in \{0, 1, \cdots\}$}
            \State $H \leftarrow \{H, \text{contour cluster}\}$ where~$\gamma_{ij} \geq 0.5$
		\EndFor
		\While{$H$ has changed}
		    \ForAll{adjacent subgraph pair~$h_i, h_j \in H$}
		        \If{$\gamma_{ij} > \Gamma$}
		            \State Merge~$h_j$ into $h_i \in H$ and delete~$h_j$ from~$H$
	            \EndIf
		    \EndFor
		\EndWhile
	\end{algorithmic} 
\end{algorithm} 

Although the MCTS-based method eventually finds an optimal solution, the running time grows rapidly with the number of contours. In this section, we present a clustering algorithm that enables reduction in the number of contour sets to consider, and allows us to speed up the MCTS substantially. At a high level, the clustering algorithm identifies a set of subgraphs of the dependency graph~$D$ called~\emph{highly dependent subgraphs} (HDS)~$H$, where~$H = \{h_1, h_2, \cdots\}$, $h_i \cap h_j = \emptyset$ and~$\bigcup_{i} h_i = C$.

Suppose we have a pair of contours such that~$(c_i, c_j) \not\in E$ that has more than one contour $c_k$ upon which they are dependent. Then $c_k$ is a \textit{dependee} and $c_i, c_j$ are \textit{dependers}. The set of common dependees of contours $c_i$ and $c_j$ is denoted as~$\phi_{ij}$ and the sets of all dependees for contours~$i$ and~$j$ are denoted as~$\phi_i$ and~$\phi_j$, respectively (i.e., $\phi_{ij} \subset \phi_i$ and~$\phi_{ij} \subset \phi_j$). Likewise, $\psi_{ij}$ and~$\psi_i$ denote the set of common dependers and set of dependers for contour~$i$. 

We can now define the \emph{degree of connectedness} $\gamma_{ij}$ as 
\begin{equation}
    \text{mean} \left( \min\left( \frac{|\phi_{ij}|}{|\phi_i|}, \frac{|\phi_{ij}|}{|\phi_j|} \right), \min\left( \frac{|\psi_{ij}|}{|\psi_i|}, \frac{|\psi_{ij}|}{|\psi_j|} \right) \right)
    .
\end{equation}
To gain intuition for this definition let us consider the circumstances under which $\gamma_{ij} = 1.$ This happens when the number of common dependers (and dependees) and single-contour dependers (and dependees) are equal. This means that $c_i$ and $c_j$ share all dependers (and dependees).

\begin{definition} [Highly connected pair]
    A pair of contours $i, j$ is \emph{highly connected} if the degree of connectedness~$\gamma_{ij}$ is greater than or equal to a constant $\Gamma$.
\end{definition}

\noindent Figure~\ref{fig:example_hds} illustrates an example of a dependency graph where each node represents a contour and an edge represents dependency where dependee nodes are placed above dependers. Contours~$i$, $j$ and~$k$ have at least one common dependees. The degree of connectedness between~$i$ and~$j$ is 0.593, where as that between~$i$ and~$k$ is only 0.25. In Sec.~\ref{sec:experiments} we show that a choice of $\Gamma=0.5$ works well in practice.

\begin{definition} [Highly dependent subgraph (HDS)] \label{def:hds}
    A subgraph~$h$ of dependency graph~$D$ is \emph{highly dependent} if all pairs of non-dependently adjacent contours with more than one common dependees or dependers (i.e., $|\phi_{ij}|>0$ or~$|\psi_{ij}|>0$ for $ij$-pair) are highly connected.
\end{definition}

\noindent When $\gamma_{ij}=1$, a pair is said to be \emph{fully connected}. A subgraph is said to be \emph{fully dependent} when all pairs of non-dependently adjacent contours are fully connected.

\begin{figure*}[ht]
	\centering
	\subfloat[Toolpath from Slicer, top view]{\includegraphics[width=0.55\columnwidth]{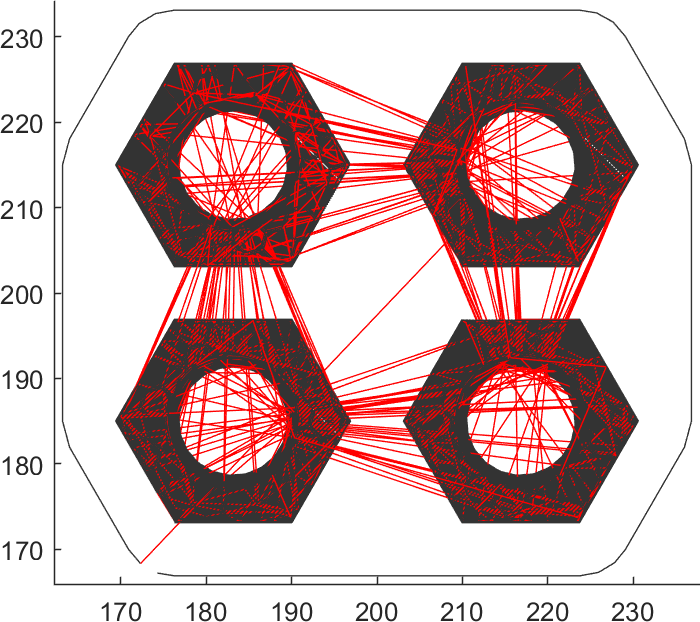} \label{fig:nuts_top_slicer}}
	\subfloat[Toolpath from local search, top view]{\includegraphics[width=0.55\columnwidth]{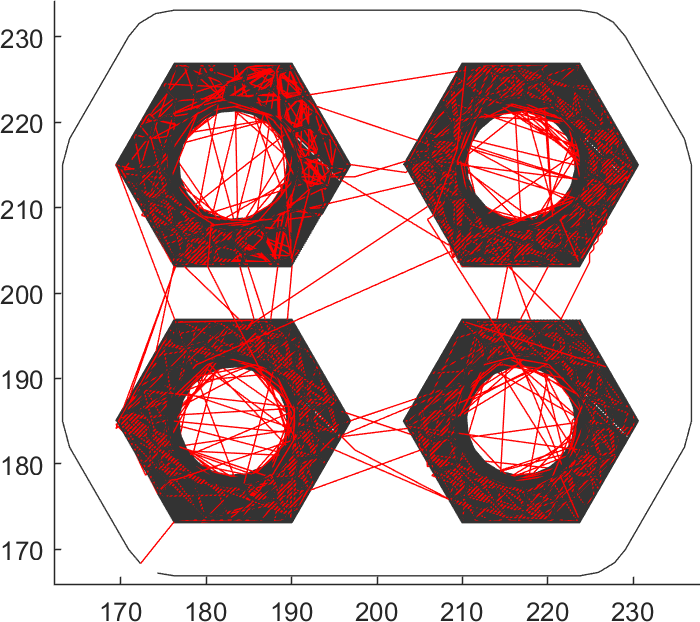} \label{fig:nuts_top_sam}}
	\subfloat[Toolpath and clusters from MCTS, top view]{\includegraphics[width=0.55\columnwidth]{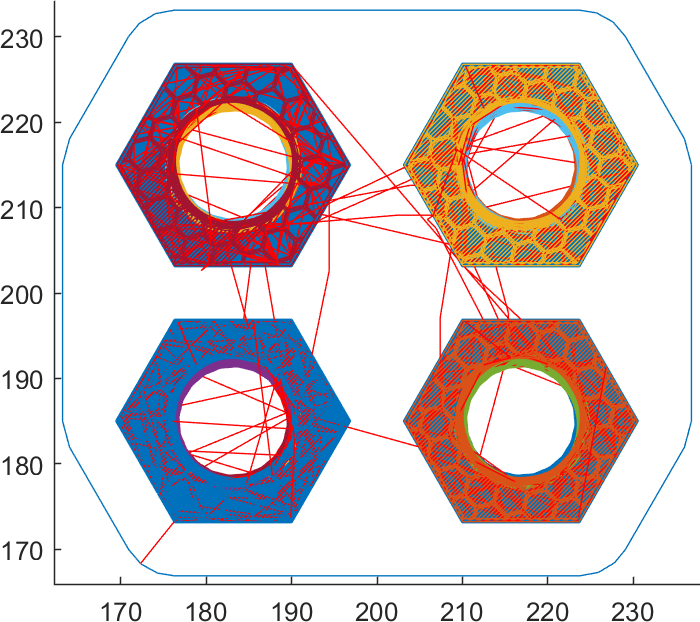} \label{fig:nuts_top_mcts}}
	\caption{`Four nuts' model. Toolpath for building the part by (a)(d) typical layerwise planner, (b)(e) local search from \cite{LensgrafMettuIROS18}, and (c)(f) proposed MCTS, with red indicating non-printing motion. Solution toolpath for each method is shown in red. Extrusionless distances (in mm) are 16737, 12220 and 11057, respectively.}
	\vspace{-4ex}
	\label{fig:nuts}
\end{figure*}
\begin{figure*}[ht]
	\centering
	\subfloat[Toolpath from Slicer]{\includegraphics[width=0.5\columnwidth]{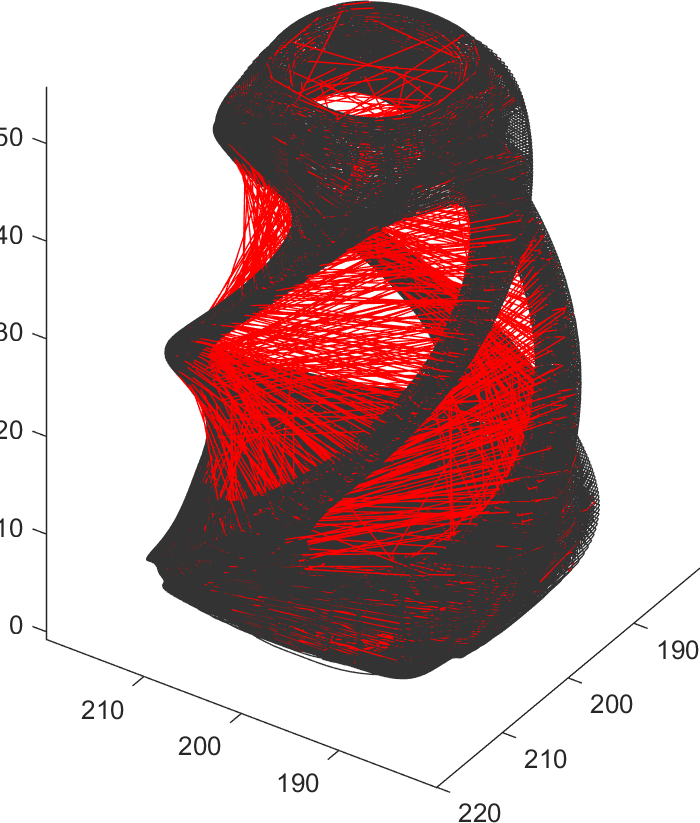} \label{fig:twisty_slicer}}
	\subfloat[Toolpath from local search]{\includegraphics[width=0.5\columnwidth]{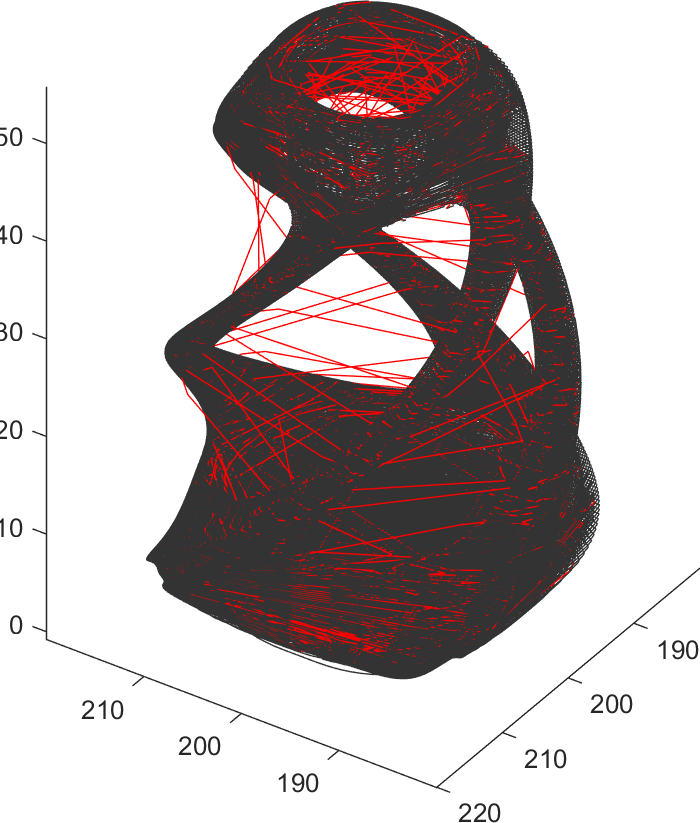} \label{fig:twisty_sam}}
	\subfloat[Toolpath and clusters from MCTS]{\includegraphics[width=0.5\columnwidth]{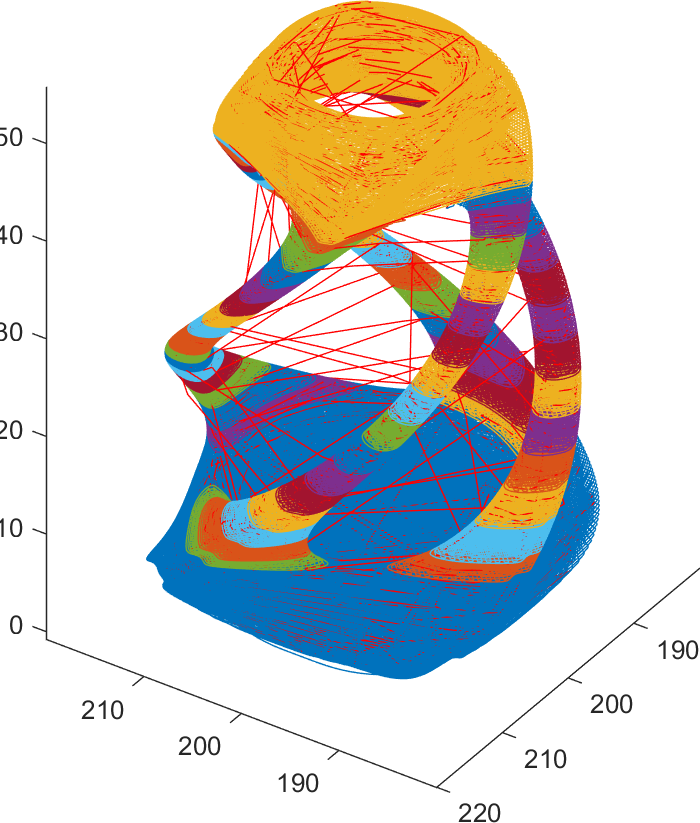} \label{fig:twisty_mcts}}
	\caption{`Twisty' model. Toolpath for building the part by (a) typical layerwise planner, (b) local search from \cite{LensgrafMettuIROS18}, and (c) proposed MCTS, with red indicating non-printing motion. Solution toolpath for each method is shown in red. Extrusionless distances (in mm) are 25021, 11423 and 11306, respectively.}
	\vspace{-4ex}
	\label{fig:twisty}
\end{figure*}
\begin{figure*}[ht]
	\centering
	\subfloat[Toolpath from Slicer]{\includegraphics[width=0.55\columnwidth]{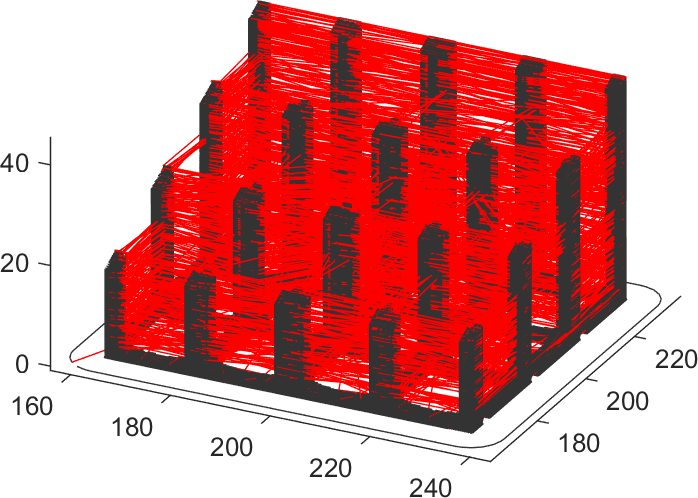} \label{fig:kitchen_slicer}}
	\subfloat[Toolpath from local search]{\includegraphics[width=0.55\columnwidth]{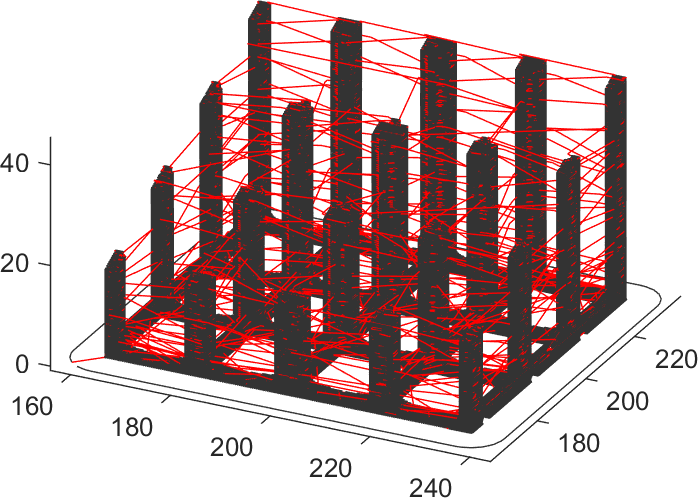} \label{fig:kitchen_sam}}
	\subfloat[Toolpath and clusters from MCTS]{\includegraphics[width=0.55\columnwidth]{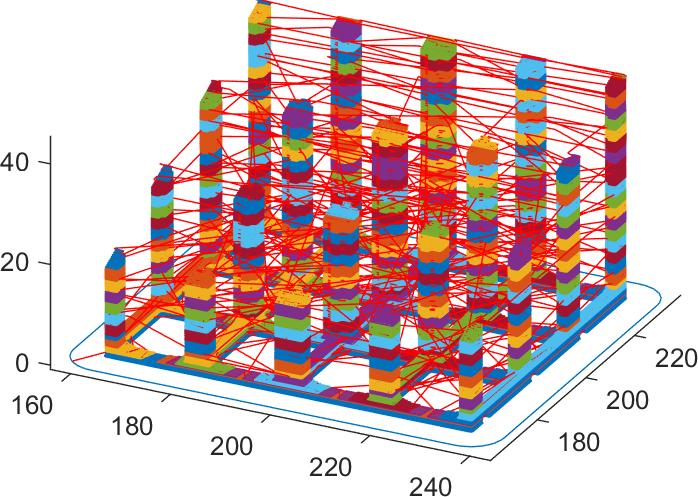} \label{fig:kitchen_mcts}}
	\caption{`Sponge holder' model. Toolpath for building the part by (a) Sli3r, (b) local search from \cite{LensgrafMettuIROS18}, and (c) MCTS. Red indicates extrusionless travel, and multiple colors in (c) indicates HDS components of the dependency graph. Extrusionless distances (in mm) are 84340, 26809 and 33665, respectively.}
	\vspace{-2ex}
	\label{fig:kitchen}
\end{figure*}

The pseudocode for finding a set of HDS is shown in Alg.~\ref{alg:clustering}. Initially, we compute the \emph{dependency depth} for all contours in dependency graph~$D$. The dependency depth is the number of edge transitions from root contours. For example, the dependency depth for contour~$i$ in Fig.~\ref{fig:example_hds} is~$1$ and $2$ for all dependees. Given all contours at same depth, we enumerate each pair~$ij$ and find subgraphs~$H$ where the degree of connectivity is greater than~$\Gamma$. For every adjacent subgraphs~$h_i \in H$ and~$h_j \in H$, we compute the degree of connectivity. If the degree is greater than~$\Gamma$, then this pair is merged. We repeat this process until no further merging is possible. We refer to the resulting dependency graph as the \emph{clustered dependency graph}. Figure~\ref{fig:example_clustering_overview} shows how contours are clustered into three different subgraphs, where Fig.~\ref{fig:example_clustered} shows the final form of the clustered dependency graph. In Fig.~\ref{fig:clustering_examples}, we demonstrate a various models with HDS.

The overall goal of clustering the dependency graph is to create subgraphs in which optimization of the toolpath beyond a layer-by-layer solution is not needed. Each subgraph in the clustered dependency graph is highly connected by construction.  Therefore, before a contour at given depth can be printed, most of its dependees must be printed. In the extreme case of a fully dependent subgraph where all contour pairs are fully connected, all dependees must be printed. This property implies that the optimal printing sequence is such that each dependee layer (i.e., lower) must be printed in full before printing the depender layer (i.e., upper). 

Such a clustering can be achieved by selecting a strict enough $\Gamma$. Once we identify a set of HDS on the original dependency graph is identified, we can then use the clustered dependency graph as input to MCTS to greatly reduce runtime. This approach is justified by the fact that the toolpath cost within an HDS cannot vary significantly and thus the reduced representation will yield a high quality toolpath. In our implementation we keep the threshold $\Gamma=0.5$ to achieve a balance of efficiency and approximation. 

\section{EXPERIMENTAL STUDIES} \label{sec:experiments}

\begin{figure}[t]
	\centering
	\includegraphics[width=0.9\columnwidth]{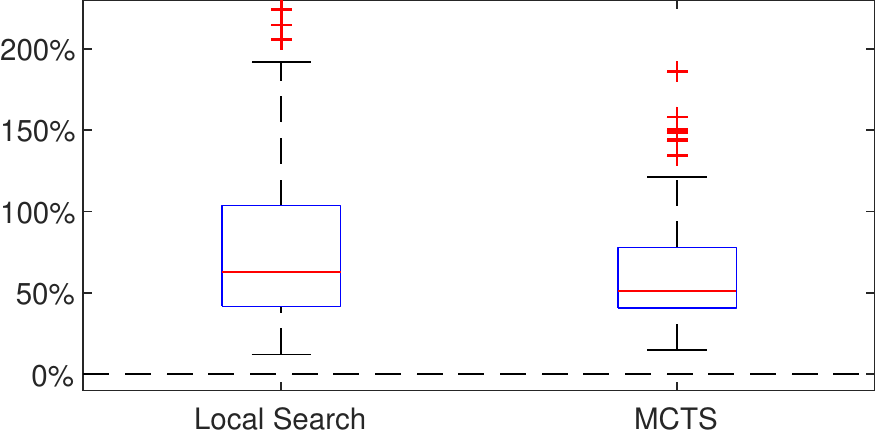}
	\caption{Reduction in extrusionless travel of local search~\cite{LensgrafMettuICRA2017} and MCTS in comparison to Slic3r.}
	\vspace{-2ex}
	\label{fig:boxplot}
\end{figure}
\begin{figure}[h]
	\centering
	\subfloat[`Four nuts']{\includegraphics[width=0.8\columnwidth]{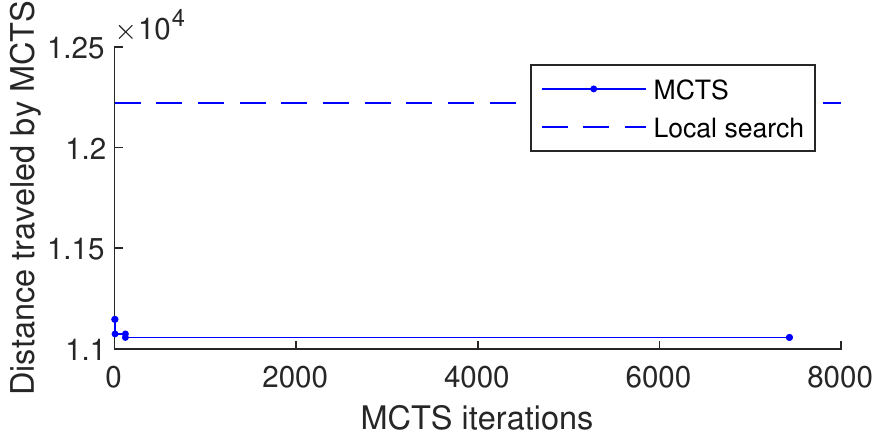} \label{fig:conv_four_nuts}}
	\vspace{-2ex}
	\\
	\subfloat['Twisty']{\includegraphics[width=0.8\columnwidth]{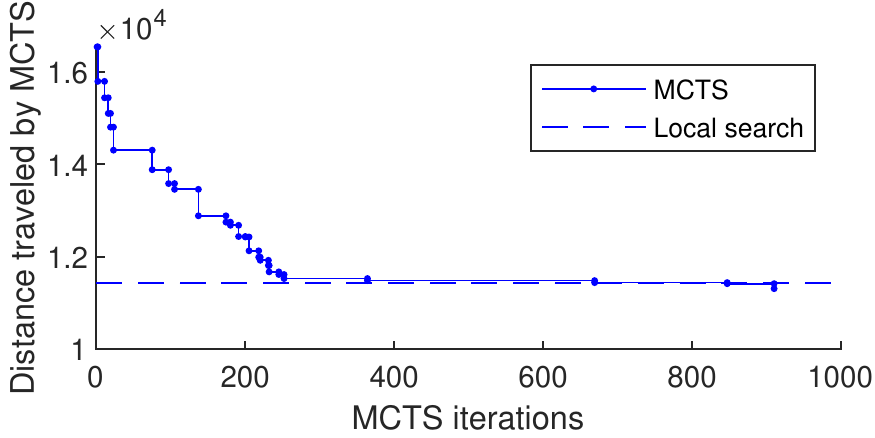} \label{fig:conv_twisty}}
	\vspace{-2ex}
	\\
	\subfloat[`Sponge holder']{\includegraphics[width=0.8\columnwidth]{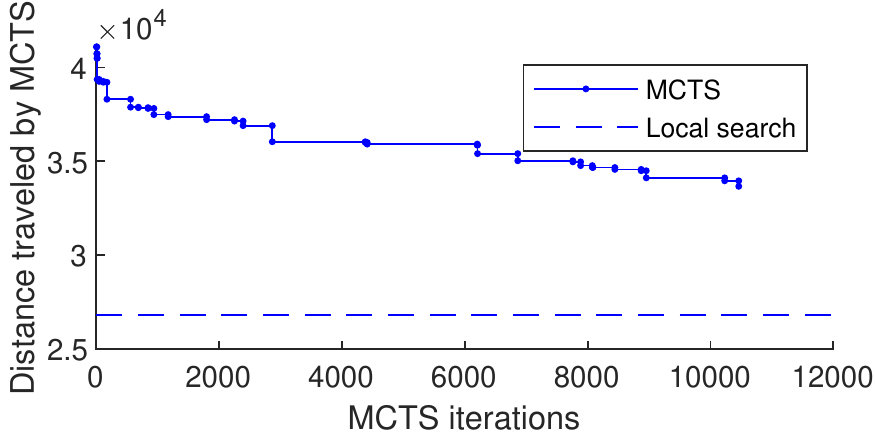} \label{fig:conv_kitchen}}
	\caption{extrusionless travel of the MCTS solution in each iteration for examplar models. Dashed line shows the cost of the local search solution.}
	\vspace{-2ex}
	\label{fig:conv}
\end{figure}
\begin{figure}[t]
	\centering
	\subfloat[`Four nuts' - MCTS performed better] {\includegraphics[width=1\columnwidth]{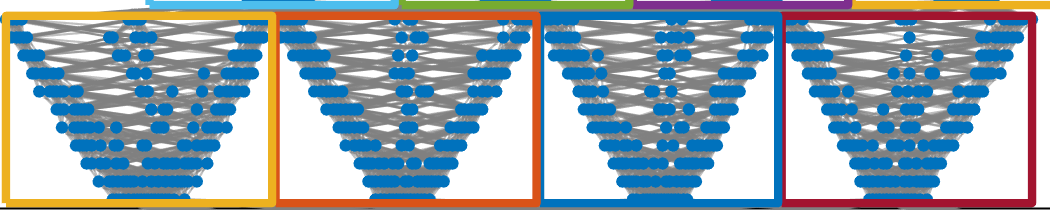} \label{fig:parts_four_nuts}}
	\vspace{-1ex}
	\\
	\subfloat['Sponge holder' - MCTS performed worse] {\includegraphics[width=1\columnwidth]{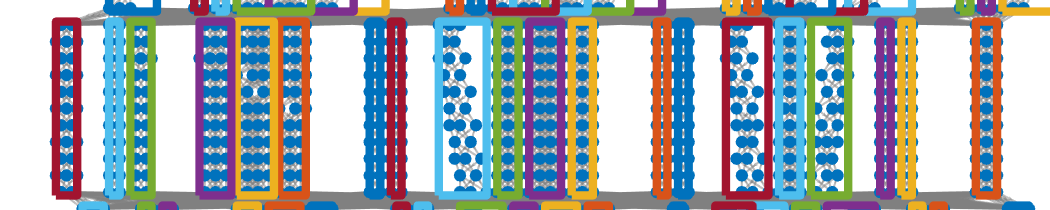} \label{fig:parts_twisty}}
	\caption{Parts of dependency graphs and clusters for two exemplar models}
	\vspace{-2ex}
	\label{fig:parts}
\end{figure}


In this section we present experimental validation of the algorithm presented in Sec.~\ref{sec:mcts}. To more closely examine the performance of MCTS against that of local search, we chose a subset of 75 models from our prior benchmark of 400+ models~\cite{LensgrafMettuICRA2017}. We implemented our MCTS algorithm in Matlab and measured the reduction in extrusionless travel relative to our existing local search algorithm and Slic3r as a baseline comparison. As shown in the boxplots in Fig.~\ref{fig:boxplot}, the median reduction of extrusionless travel for local search is about 60\% and 55\% for MCTS. The running times of our MCTS algorithm were on the order of around 6 minutes on average; MCTS was terminated if the solution did not improve over 5 minutes of search.

While local search is guaranteed only to reach a locally optimal solution, MCTS is guaranteed to converge asymptotically to the global optimum. In our implementation, we terminated the MCTS search when there is no reduction in solution cost after 5 minutes. 

As noted above, MCTS performs about as well as local search. Indeed, we noted in prior work~\cite{LensgrafMettuICRA2017} that it seemed difficult to eke any more improvement from local search -- even with an ideal extruder geometry (i.e., no constraints on reachability), extrusionless travel could only be reduced an additional few percent relative to Slic3r. We were able to also show that at least for a few small models, local search achieved the LP-relaxation lower bound for a linear program set up to find an optimal toolpath in the dependency graph. 

We view the MCTS results as an empirical extension of this reasoning. Since MCTS guarantees eventual convergence to optimal, we cannot say with certainty that our solutions are optimal. Indeed, some MCTS toolpaths are slightly poorer quality than those produced by local search. To study performance of MCTS more closely, we partitioned the toolpaths into three categories corresponding to whether they were  better (over 10\% improved), about the same (within 5\%), or worse (over 10\% worsen), than the local search toolpaths.
To look at concrete examples of these categories, we chose exemplar models for each of these categories. Figures~\ref{fig:nuts}, \ref{fig:twisty} and~\ref{fig:kitchen} give toolpaths for the chosen model for each category, respectively, for Sli3r, local search, and MCTS. In Fig.~\ref{fig:conv}, we show MCTS convergence trends for three different models in relation to the local search solution cost. 

In `four nuts' (Fig.~\ref{fig:conv_four_nuts}), extrusionless travel rapidly reduces in the early iterations and quickly outperforms the local search solution. In `twisty' (Fig.~\ref{fig:twisty}), the MCTS solution starts off worse than the local search and surpasses the local search solution cost after about $8000$ iterations. In `sponge holder' ((Fig.~\ref{fig:kitchen}), the distance reduced gradually but reached a point of nonimprovement and then timed out after 5 minutes.

To gain insight into the relative behavior of MCTS and local search, in Fig.~\ref{fig:parts} we show the parts of dependency graph and their HDS components for `four nuts' and `sponge holder.' In `four nuts', the number of contours within a HDS was much higher than in `sponge holder'. Thus the size of the clustered dependency graph given as input to MCTS is smaller and results in rapid convergence. In contrast, there exist many more HDS components in `sponge holder.' This implies that the convergence rate for `sponge holder' is likely much slower, with an early termination resulting in a poor solution. One way to think about the relative performance of these models is in terms of clustering threshold parameter $\Gamma$. While we set the value of $\Gamma=0.5$ to balance solution quality and efficiency, the best value is in fact model dependent and for models such as 'sponge holder' a smaller value would have resulted in a more tractable dependency graph.

The discussion implies that the solution approaches true optimal as the number of clusters increases (i.e., running MCTS over a set of contours, not clusters). However the convergence rate would become poorer. On the other hand, we would find a converging solution fast with less clusters, but the solution could be further from true optimal if we may violate our assumption in Def.~\ref{def:hds} by forcing high~$\Gamma$. Finding the right~$\Gamma$ to balance remains as open question.





\section{DISCUSSION}

In this paper we have developed the first provably convergent algorithm for FDM toolpath planning. In order for our Monte Carlo Tree Search approach to be feasible, we also developed a novel clustering approach to compress the dependency graph of the input model. We tested our algorithm over 75 models an observed similar overall performance on average, but empirically characterized the behavior of MCTS when it appears to converge. 

A natural question is why one would use MCTS over local search for a given model. Using our empirical studies, it appears that the output of the clustering step and subsequent composition of HDS components of the dependency graph provide guidance as to whether MCTS can achieve convergence. As we saw in our empirical analysis if there enough HDS components with respect to the size of the dependency graph then it is highly likely that MCTS will converge to an optimal toolpath. If the number of HDS components is too large, or the average size is too small, then MCTS will have difficulty exploring the toolpath space and may perform worse than local search. 



\newpage
\balance
\bibliographystyle{IEEEtran}
\bibliography{references}

\end{document}